\documentclass[conference]{IEEEtran}
\IEEEoverridecommandlockouts
\usepackage{cite}
\usepackage{multirow}
\usepackage{amsmath,amssymb,amsfonts}
\usepackage{algorithmic}
\usepackage{graphicx}
\usepackage{textcomp}
\usepackage{xcolor}

\def\BibTeX{{\rm B\kern-.05em{\sc i\kern-.025em b}\kern-.08em
    T\kern-.1667em\lower.7ex\hbox{E}\kern-.125emX}}

\title{Toward Accurate Platform-Aware Performance Modeling for Deep Neural Networks}

\makeatletter
\newcommand{\linebreakand}{%
  \end{@IEEEauthorhalign}
  \hfill\mbox{}\par
  \mbox{}\hfill\begin{@IEEEauthorhalign}
}
\makeatother

\author{\IEEEauthorblockN{Chuan-Chi Wang}
\IEEEauthorblockA{
\textit{ADLINK Technology Inc.}\\
New Taipei, Taiwan \\
chuan-chi.wang@adlinktech.com}
\and
\IEEEauthorblockN{Ying-Chiao Liao}
\IEEEauthorblockA{
\textit{National Taiwan University}\\
Taipei, Taiwan \\
d08922004@ntu.edu.tw}
\and
\IEEEauthorblockN{Ming-Chang Kao}
\IEEEauthorblockA{
\textit{ADLINK Technology Inc.}\\
New Taipei, Taiwan \\
fencer.kao@adlinktech.com}
\linebreakand
\IEEEauthorblockN{Wen-Yew Liang}
\IEEEauthorblockA{
\textit{ADLINK Technology Inc.}\\
New Taipei, Taiwan \\
william.liang@adlinktech.com}
\and
\IEEEauthorblockN{Shih-Hao Hung}
\IEEEauthorblockA{
\textit{National Taiwan University}\\
Taipei, Taiwan \\
hungsh@csie.ntu.edu.tw}
}

\begin{document}

\maketitle

\begin{abstract}
In this paper, we provide a fine-grain machine learning-based method, \emph{PerfNetV2}, which improves the accuracy of our previous work for modeling the neural network performance on a variety of GPU accelerators. Given an application, the proposed method can be used to predict the inference time and training time of the convolutional neural networks used in the application, which enables the system developer to optimize the performance by choosing the neural networks and/or incorporating the hardware accelerators to deliver satisfactory results in time. 
Furthermore, the proposed method is capable of predicting the performance of an unseen or non-existing device, e.g. a new GPU which has a higher operating frequency with less processor cores, but more memory capacity. This allows a system developer to quickly search the hardware design space and/or fine-tune the system configuration.
Compared to the previous works, \emph{PerfNetV2} delivers more accurate results by modeling detailed host-accelerator interactions in executing the full neural networks and improving the architecture of the machine learning model used in the predictor. Our case studies show that \emph{PerfNetV2} yields a mean absolute percentage error within 13.1\% on LeNet~\cite{lenet}, AlexNet~\cite{alexnet}, and VGG16~\cite{vgg16} on NVIDIA GTX-1080Ti, while the error rate on a previous work \cite{daniel2018} published in ICBD 2018 could be as large as 200\%.
\end{abstract}


\section{Introduction}

Deep learning (DL) has explosive growth and gradually deployed for many application scenarios in enterprise domains, such as email searching, face recognition, human detection, autonomous driving, etc. \cite{ai_email_search, ai_face_recognition, ai_human_detection, ai_autonomous_driving}. 
In recent researches, DL models become deeper and more complex to get the better accuracy \cite{resnet, vgg16, inceptionv3}. However, most of the time, these large DL models may not meet the timing constraints or memory requirements on the hardware, especially on the edge devices. In light of this, many DL researchers focus on designing the efficient DL models under various hardware limitations \cite{yolov3, mobilenet}. In a general case, DL models is trained on powerful workstations, but would be deployed on a edge device with constrained resources. If the DL model designer underestimates the inference time on the edge device, then the DL model would not be able to meet the requirement in a real-time application.
Moreover, deep learning accelerators (DLAs) are making their way into edge devices. 
The developer of a DL application on an edge device may find that it is complex to design the DL model and choose the DLA at the same time, as it is tedious and sometimes impossible to set up an actual system to evaluate the performance of the DL application.

\begin{figure}
\centering
\includegraphics[width=3.5in]{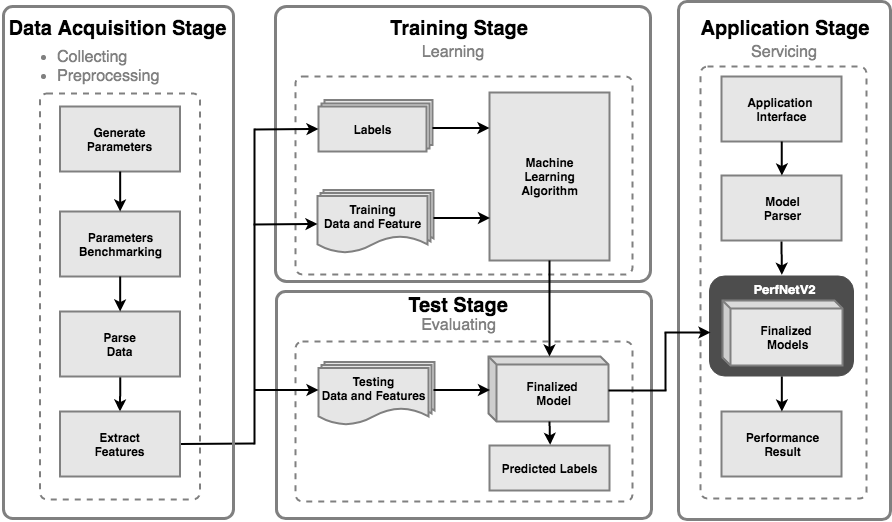}
\caption{Overview of the performance prediction service} 
\label {fig:application_workflow}
\end{figure}

The aforementioned hardware/software co-design problem for DL is increasingly important for efficiently processing high-volume, high-velocity big data coming from Internet of Things (IoT). 
Thus, we hope to establish an interactive service to simplify the performance evaluation and optimization process for the edge-based DL application developers by integrating a set tools, as shown in Figure \ref{fig:application_workflow}, which perform automatic data acquisition, train the prediction model based on the acquired data, evaluate the prediction model via testing, and provide the performance prediction service to the application developer, as the following:
\begin{enumerate}

\item 
\emph{Data Acquisition Stage:} We provide tools to collect the data regarding to how the developer's system performs for convolutional neural networks. Since the design space for convolutional neural networks is extremely large, one cannot collect the performance data by brute-force. In \cite{daniel2018}, a sampling method was proposed to acquire a representative data set by measure the elapsed time to complete a neural network layer (i.e. convolutional, max pooling, dense) with respect to selected batch sizes, matrix sizes, kernel sizes, etc.
However, as we further experiment with this approach, we see that the prediction accuracy increase with the data set, but at the cost of data collection time and the training time in the Training Stage. 
Furthermore, we found that simply measuring the elapsed time to complete a neural network layer is far too coarse to establish an accurate performance model, as the execution involves complicated interactions between the host (CPU) and the accelerator (GPU) in an actual system.


\item 
\emph{Training Stage:} In this stage, a machine learning is trained with the representative sample data set acquired in the \emph{Data Acquiring Stage}. The basic architecture and loss function of our machine learning model is inspired from Daniel et. al. \cite{daniel2018}. While we improve the data acquisition method to additionally consider the CPU-GPU interactions, we also enhance the machine learning model by adding several layers to our previous model and adopt the mean absolute percentage logarithmic error as the loss function. The changes result in significant improvement in the prediction accuracy.


\item 
\emph{Test Stage:} We choose 4 different metrics: MAE, MAPE, RMSE, and $R^2$, which are described in details in Section \ref{subsection:hard_devices}, to measure the accuracy of the produced model during this stage. 


\item 
\emph{Application Stage:} After a model shows acceptable accuracy during the test, it becomes available for the users via a set of application programming interfaces, which enables
the users to upload their deep neural network architectures to our on-line service and returns the predicted performance results for a list of DLA devices.
This allows the users to select suitable neural network architectures and DLA's based on the application requirements. 

\end{enumerate}


In this paper, we further propose platform-aware performance modeling techniques to improve the accuracy of prediction and provide extra functionalities. 
The contributions of this paper are briefly described as follows:

\begin{itemize}
\item A novel performance predict service is proposed to significantly increase the productivity of system developers and support hardware/software co-design projects.

\item A simple, yet effective formula for estimating total neural network training time is proposed.

\item A neural network structure with convolutional layers, is proposed to increase the accuracy for predicting both inference and training time.

\item The performance impact due to hardware changes can be predicted with the proposed approach to assist system designers.
\end{itemize}

The remaining of this paper is organized as the following. Section 2 mentions the the related works. 
Section \ref{section:methodology} described the proposed methods. 
Section \ref{section:data_collection} further describes the actual training process to produce a practically accurate model within a reasonable time span. 
Section \ref{section:performance_evaluation} evaluates the effectiveness of the produced model via experimental results. 
Section \ref{section:conclusion_futurework} concludes the paper and discusses potential future works. 

\section{Related Work}
Recently, convolutional neural network (CNN) replace traditional feature extraction algorithms and are able to outperform general machine learning models in terms of accuracy for many classification problems with large training data.
The architecture of the neural network and the configuration of the hyper-parameters would heavily affect a deep learning (DL) model accuracy.
In 2015, the depth of some DL models such as ResNet has already reached 200 and the depth keeps growing recently to improve the accuracy, but not considers the cost of computing power.
In the meanwhile, instead of using CPU to perform the computation in a DL model, many DL applications are running on heterogeneous systems which utilize graphics processing unit (GPU) or deep learning accelerators (DLA's) to carry out the DL computations, which significantly reduces the elapsed time and/or throughput of DL tasks. 
However, the resulted elapsed time and throughput may depend heavily on the architecture of such special purpose processors and cannot be estimated simply based on the number of computing operations generated by a DL model.

The traditional benchmark tool, such as passmark \cite{passmark} providing the scoring system representing the performance to understand the performance of special-purpose processors.
This performance score may not fit for DL problems, because it integrated too many system considerations rather than DL's parallel vector operation. Therefore, several research teams proposed to build benchmarks for DL, e.g. DAWNbench \cite{dawnbench} and MLPerf \cite{mlperf}. 
While the basic performance metrics such as throughput (operations/sec) and latency of these benchmark tools for performing an inference job, they cannot be used to accurately predict the performance of any DL tasks, as they only cover a small design space of neural network models, whose characteristics may be very different from the DL task in real scenarios.

To evaluate the inference time of DL models, Qi et. al. provided an open-source tool called PALEO~ \cite{paleo2017}, which proposed a method to construct an analytical performance model. 
PALEO took a small number of representative DL workloads which run operations for a short time on a single GPU to estimate a platform-dependent parameter called \emph{platform percent of peak} (PPP) which captures the average relative inefficiency of the platform compared to peak FLOPS. However, as Daniel et. al.~\cite{daniel2018} pointed out the analytic model did not consider the architectural details and simply assumed that the execution time of a neural network scaled linearly with the number of operations, which could produce inaccurate results.

An alternative approach, which Daniel et. al. \cite{daniel2018}  proposed, utilizing several trained DL models for each target platform based on measured data. They figured out a neural network as a combination of individual layers so that their approach was capable of predicting the execution time of each individual layer and then connect all individually predicted result to estimate the total execution time of the full neural network model. Unfortunately, while the method gives more accurate results than PALEO does for estimating the total execution time on CPU platforms, it does not provide accurate estimates for GPU platforms in our experimental environments, which is considered as a coarse-grained method.

In order to attain the fine-grained data, we need to use a number of performance analysis tools such as \emph{SOFA}\cite{sofa2018}, \emph{TensorFlow Profiler}\cite{tfprofiler}, etc. to profile the computational time in a heterogeneous computing system and the data transfer time between CPU and DLA.
The profile information includes not only collects function call traces to reveal the workflow and break down the execution time for a DL task but the basic CPU and GPU utilization. 
Chuan-Chi et. al. proposed \emph{PerfNet} \cite{perfnet} which using the profiling tool split layers into three phases, and designed a multi-layer regression models to predict a more accurate inference time than Daniel et. al. does. In order to improve the accuracy of the predicted results compared with the previous works, we based on \emph{PerfNet} and proposed a new neural network structure with a specialized loss function.

\section{Methodology}
\label{section:methodology}

The performance of a DL application depends heavily on the system configuration. However, it is difficult to model and predict the application performance with an analytic method because it involves complex hardware-software interactions. While previous works such as \cite{daniel2018} attempted to establish a performance model for DL applications via machine learning, it requires an enormous amount of data by measuring the execution time of various workloads. Unfortunately, the simple data-driven machine learning approach yields inaccurate results sometimes, especially for the case of systems with GPUs, since the approach did not model the interactions between the host and the GPU at all.

On the other hand, when we measure the elapsed time for each neural network layer, we can include the communication time required for the host to transfer the input data to the GPU. For example, Table \ref{tab:LeNet_real_time} lists the elapsed time for each layer in LeNet-5, which is individually measured on a system with NVIDIA-1080Ti and a batch size is equal to 1. If we naively calculate the sum of these individually measured elapsed time to predict the forward propagation time of LeNet-5, then the result, 3.231ms, is much higher than the actual elapsed time, 0.563ms. Since the internal layers receive their input from their previous layers within the same GPU, they do not require data transfer from the host to the GPU.  
In PerfNet \cite{perfnet}, we propose to build a performance model for inference time on GPU-based systems by additionally considering the interactions between the host and the GPU, as well as the order of the layers. In this work, we enhance the previous work with support of training time. The modeling workflow is discussed in Section \ref{section:time_modeling}. The procedures for predicting the inference time and training time are described respectively in Section \ref{section:inference_time_procedure} and  \ref{section:training_time_procedure}. 

To further improve the prediction accuracy, 
we propose two specialized loss functions and two network structures.
The combination of \emph{mean absolute percentage logarithmic error} (MAPLE) and a convolutional neural network delivers more accurate results for inference and training time.
Interestingly, in the case of predicting the performance for hardware changes, the combination of \emph{mean square logarithmic error} (MSLE) and use multi-layer regression delivers most accurate results. 
Section \ref{section:loss_function} and \ref{section:regression_model} describe these loss functions and neural network architectures.

\begin{table}[]
\begin{center}
\caption{{\upshape LeNet-5 on NVIDIA-1080Ti with batch size 1: Comparison of (a) the summation of all elapsed time of each layer and (b) the real elapsed time.}}
\label{tab:LeNet_real_time}
\begin{tabular}{cccc}
Name  & Input    & Operator         & Elapsed Time (ms) \\ \hline
Conv1 & 32x32x1  & convolution      & 0.523        \\
Pool1 & 28x28x6  & pooling          & 0.436        \\
Conv2 & 14x14x6  & convolution      & 0.478        \\
Pool2 & 10x10x16 & pooling          & 0.418        \\
Fc1   & 5x5x16   & dense            & 0.442        \\
Fc2   & 120      & dense            & 0.472        \\
Out   & 84       & dense            & 0.462        \\ \hline
\multicolumn{2}{l}{(a) The sum of all elapsed time} && 3.231        \\
\multicolumn{2}{l}{(b) The actual elapsed time}  && 0.563
\end{tabular}
\end{center}
\end{table}

\subsection{Workflow Model}
\label{section:time_modeling}
\begin{figure}
\centering
\includegraphics[width=3.5in]{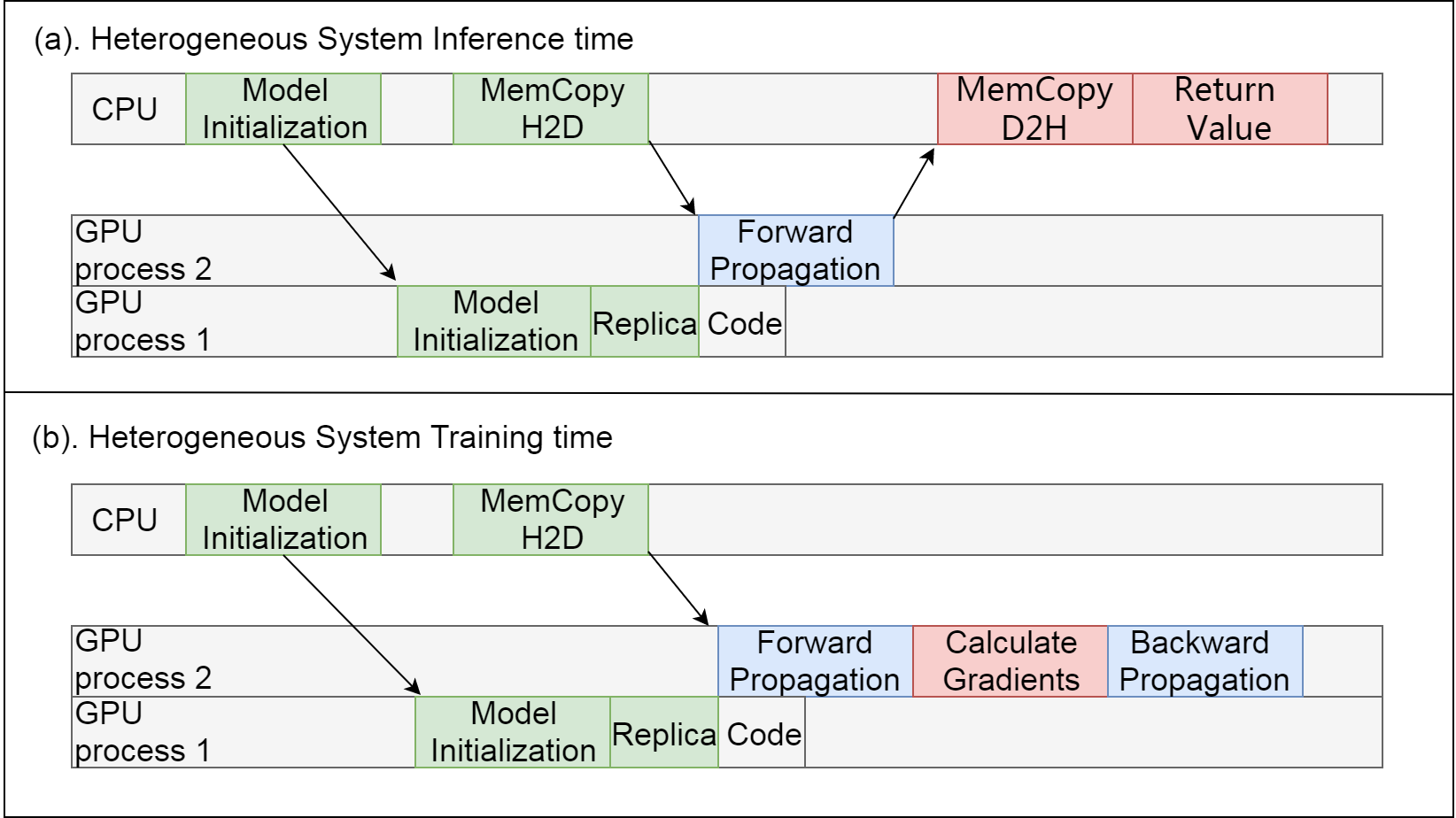}
\caption{The high-level view of the Tensorflow workflow on the heterogeneous system for (a) Inference (b) Training.}
\label {fig:timeline}
\end{figure}

In order to model the elapsed time of the full neural network on a heterogeneous system in detail, hardware researchers commonly use the Two-Phase approach \cite{paleo2017} to model the performance of the process on the heterogeneous systems. 
Two-Phase approach decomposes the total execution time into communication time and computation time.
The communication time represent the time interval of transferring data between host and device. The computation time is the length of time required to perform a computational operations on the device. 
However, in the case of modeling the neural network, the input and output of the communication task rely on the first layer architecture and the last one with different performance efficiency. With only one parameter to represent these two-step time is unreasonable and unconvincing.
It is necessary to add more parameters to cope with this problem. 
We adopted the Three-Phase approach including preprocess, execution, and postpreocess phase from \emph{PerfNet} which is concise and practical compared to another complex method.
Besides, we extend it to evaluate the performance of the training time, shown in Figure \ref{fig:timeline}. 
In order to define this tenable and easy-to-use phase for performance prediction, we breakdown the main time intervals with TensorFlow API on DLA through profiling tool including:
\begin{itemize}
\item
{\bfseries Model Initialization:} The time interval for processes which asynchronously set up the model arguments and variables on the host and device.

\item
{\bfseries MemCopyH2D:} The time interval for the the batch of the images or features transfer from host to device.

\item
{\bfseries Replica Code:} As the time interval to put the TensorFlow codes to the device queue.

\item
{\bfseries Forward Propagation:} 
The time interval for executing the operations of the layer. The operations will be launched while the data and the TensorFlow codes are both ready on DLA.

\item
{\bfseries Backward Propagation:} 
The time interval for applying the calculated gradients back to the last layer.

\item
{\bfseries MemCopyD2H:} 
The consumed time to return the inference result from GPU device to CPU host.

\item
{\bfseries Calculate Gradients:} 
The time interval for minimizing the loss value.

\item
{\bfseries Return Value:}
The time duration for returning the result from the TensorFlow framework to the user interface.
\end{itemize}

\label{section:regression_model}

\subsection{Inference Time Procedure}
\label{section:inference_time_procedure}
For the inference time jobs, in order to simplify the complexity, our neural network architecture is assumed to consist of \emph{k} layers, and the data size utilized in one iteration for the DL model is \emph{m}. we define (1) $T_{pre_{i}}$ as the time interval before GPU executes \emph{Forward Propagation} as preprocessing time for each individual layer, e.g., the total time interval of the green block shown in Figure \ref{fig:timeline}(a). (2) $T_{exe_{i}}$ is defined as the forward propagation time for individual layer in DLA, and (3) $T_{post_{i}}$ to indicate sum of the memory copy time from the GPU device to host ($T_{memCopyD2H_{i}}$) and the time to return result ($T_{retVal_{i}}$) for single layer, as follows: 

\begin{equation}
\label{post}
T_{post_{i}} = T_{memCopyD2H_{i}} + T_{retVal_{i}}
\end{equation}

Then, The inference time on heterogeneous computing system for a single batch DL model, $T_{single\_batch}$ for short, can be expressed in Equation \ref{single_batch} below:
\begin{equation}
\label {single_batch}
T_{single\_batch} = 
T_{pre_{0}} + \sum_{i=0}^{k}  T_{exe_{i}} + T_{post_{k}}
\end{equation}

In order to compute the total execution time for a single epoch of the DL model, $T_{single\_epoch(n,m)}$, defined as follow, it can be calculated with Equation \ref{total_time}, where \emph{n} is the total data size of the DL task. The total numbers of the batch which DL requires to process data are equal to \emph{n} divided by \emph{m}. 
\begin{equation}
\label {total_time}
T_{single\_epoch}(n,m) = n/m * T_{single\_batch}
\end{equation}

\begin{figure}
\centering
\includegraphics[width=3.5in]{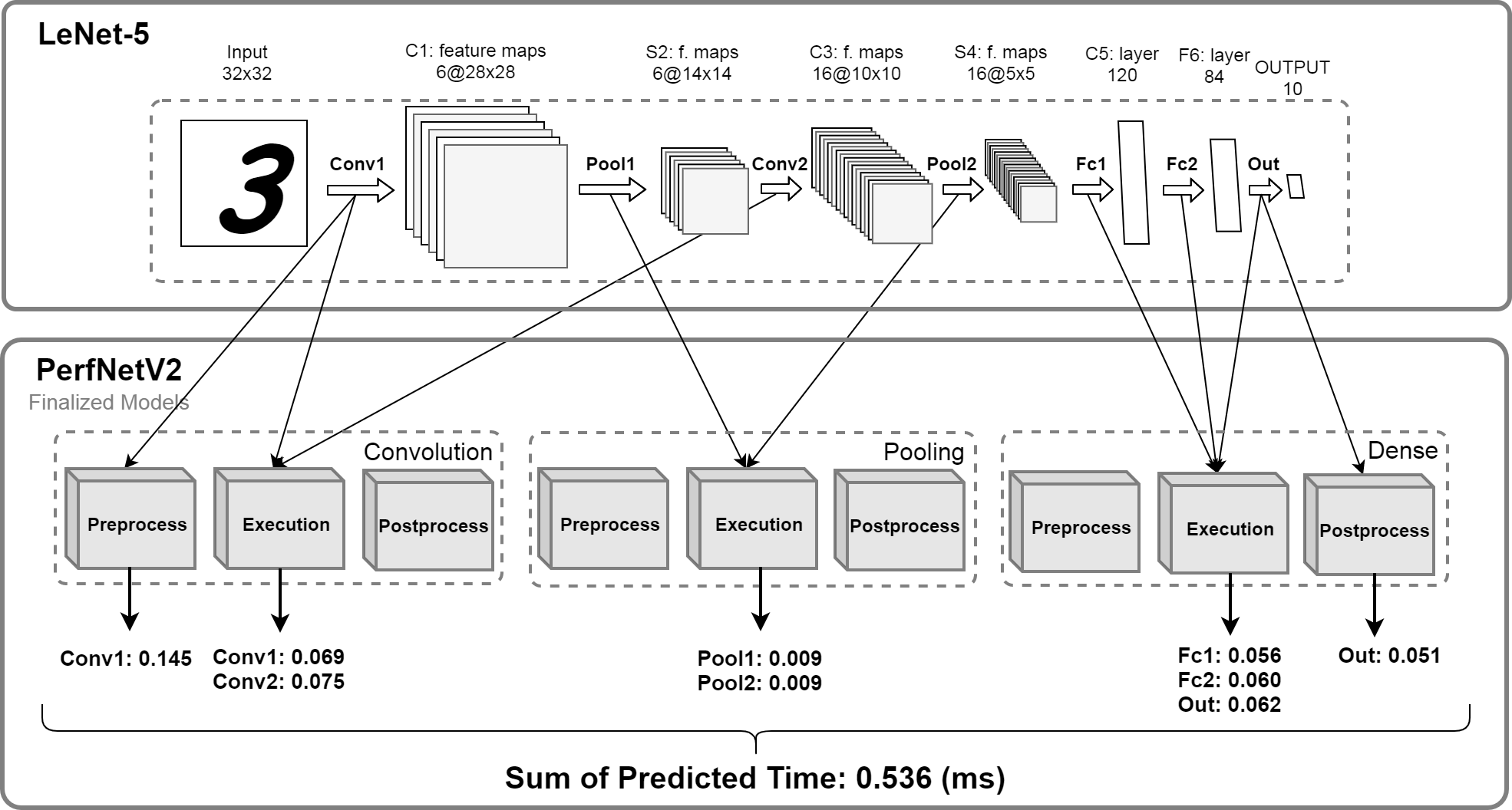}
\caption{LeNet-5 on NVIDIA-1080Ti with batch size 1: Inference Time Prediction by Our Proposed Method.}
\label {fig:predict_inference_time}
\end{figure}

  Figure \ref{fig:predict_inference_time} shows a practical example for predicting the LeNet-5 model on NVIDIA GTX 1080Ti where batch size is equal to 1 with our Three-Phase approach.
We can get predicted results Conv1: 0.145 ms from convolutional preprocess phase and Conv1: 0.069 ms from the convolutional execution phase, respectively.
Because the middle layers of the model don't need any data transmission, we only need to predict the execution phase of them. 
For the last layer, we should consider not only execution phase, but postprocess phase which represents the time interval for returning the calculated result from device to user interface, e.g., Out: 0.051 ms. We lastly add up all the predicted time, which is 0.536 ms, as our predicted result for the LeNet-5 model. Our predicted result is very close to the real actual time equal to 0.563 ms as Table \ref{tab:LeNet_real_time}, and the relative error is only less than 5\%. 

\subsection{Training Time Procedure}
\label{section:training_time_procedure}
Similarly, for training time jobs, we use the same assumptions as the inference one. \emph{k} is the total layers of the network, \emph{m} is the batch size utilized in one iteration, and \emph{n} is the total data size of the DL task. we define (1) $T_{pre}$ to represent the time interval before GPU forward propagation time, as the definition of the inference jobs. (2) $T_{post}$ as the time interval for calculating the gradient of the model. (3) $T_{exe}$ as the summation of time interval including:  \emph{Forward Propagation} and  \emph{Backward Propagation} as follow:
\begin{equation}
\label{train_exe}
T_{exe_{i}} = T_{forward_{i}} + T_{backward_{i}}
\end{equation}
With fine-tuning definition, we can use the same equation as mentioned earlier for training job. The total DL execution time of a single batch ($T_{single\_batch}$) is expressed in Equation \ref{single_batch}. The total DL execution time of single epoch $T_{single\_epoch}$ can also be expressed by Equation \ref{total_time}.

\subsection{Specific Loss Functions}
\label{section:loss_function}
In order to have a well-fitted result on the performance data set with unbalanced distribution and some noise by actual measurement, we decided to use mean absolute percentage logarithmic error (MAPLE), the same loss function as \emph{PerNet} as shown in Equation \ref{MAPLE}, for predicting inference and training time of our experimental devices. Compared to the mean square logarithmic error (MSLE) in Equation \ref{MSLE}, MAPLE intuitively decreases the effect from the outliers by the absolute operation and takes a balanced weights in each epoch by the percentage operation. In contrast to the predicted results of individual GPU cards, we use MSLE on unseen hardware prediction. The distribution of unseen hardware jobs is quite uncommon and entirely different from individual case so that MAPLE does not work well in this training process.

\begin{equation}
\label{MAPLE}
MAPLE = \frac{1}{n} \sum_{i=0}^{n}
 \left |  \frac{log(1+\hat{y_{i}}) - log(1+y_{i})}{log(1+y_{i})}\right |
\end{equation}

\begin{equation}
\label{MSLE}
MSLE = \frac{1}{n} \sum_{i=0}^{n}
 [log(1+\hat{y_{i}}) - log(1+y_{i})] ^{2}
\end{equation}

\subsection{Convolutional Regression Network}
\label{section:regression_model}
\begin{figure}
\centering
\includegraphics[width=2.8in]{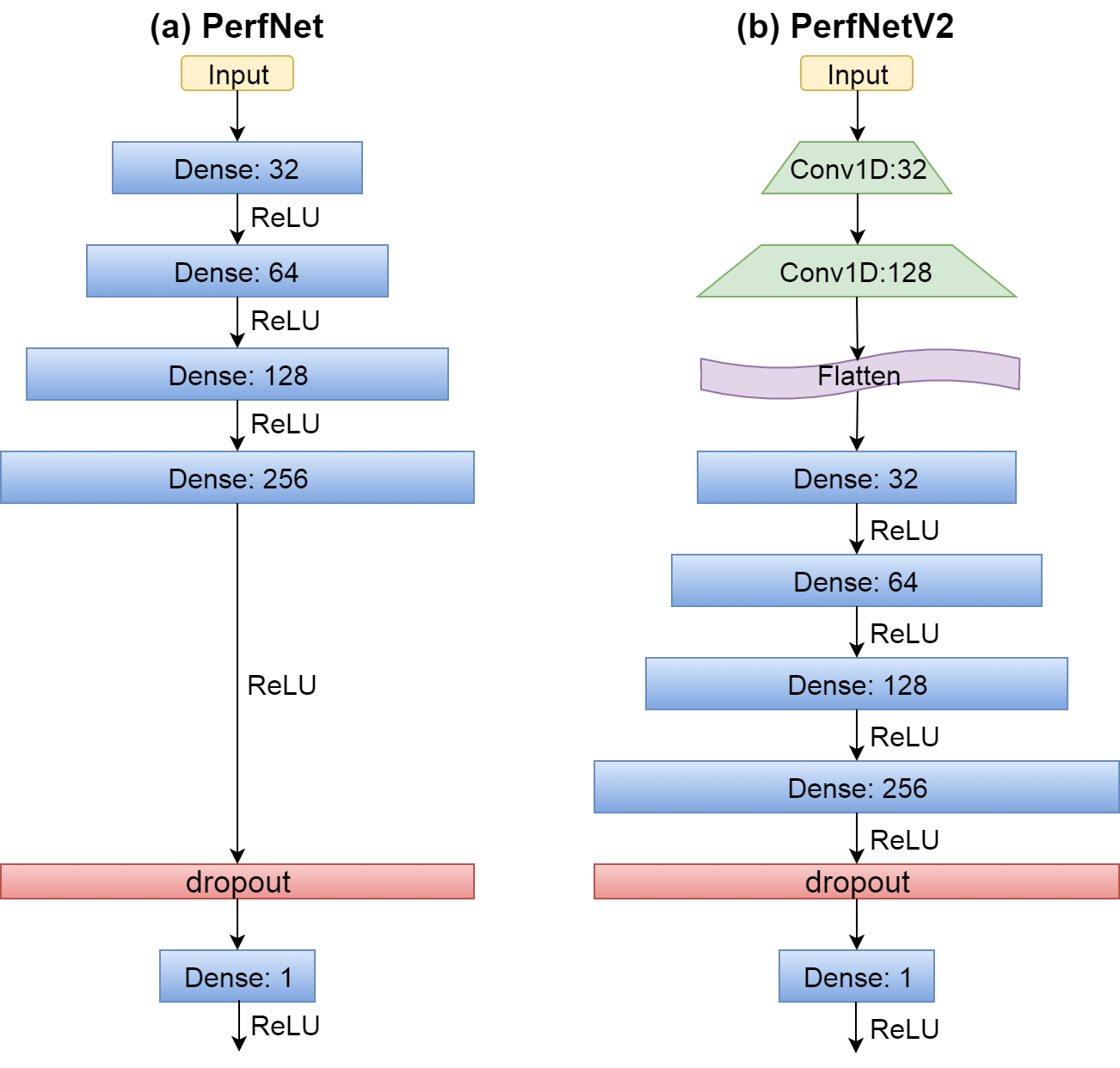}
\caption{Comparsion of (a) PerfNet architecture constructed by multi-layer regressions and (b) PerfNetV2 architecture. The green blocks are one-dimension convolutional layers and the blue blocks are fully connection layers. The output size of the layer is shown above each block.}
\label {fig:net_architecture}
\end{figure}

To predict the three phases time for the inference/training time intervals of each layer, i.e. $T_{pre_{i}}$, $T_{exe_{i}}$, and $T_{post_{i}}$ in Equation \ref{single_batch}, we propose a convolutional neural network, named as \emph{PerNetV2}, rather than the previous mlti-layer regression models in previous works \cite{daniel2018, perfnet}.
In \emph{PerNet} network architecture analysis, the over-fitting may occur while layer is deeper than 4 with unbalanced data.
In order to prevent over-fitting, we add two convolutional layers in front of \emph{PerNet} architecture as the feature extractor to reduce the phenomena, as shown in Figure \ref{fig:net_architecture}.
The first convolutional layer filters the input features with 32 kernels of size 3 x 1 with a stride of 1.
The second convolutional layer takes the output of the first convolutional layer as input and filters it with 128 kernels of size 2 x 1 with stride of 1. 
The third layer is called flatten layer, used to flatten the input for the following dense layer.
The fourth to eighth layers are fully-connected layers with ReLu function, and the output of neurons is from 32 to 256 which is the same as \emph{PerfNet}.
The ninth layer is the dropout layer with 0.3 dropout rate to avoid over-fitting. 
The last layer of \emph{PerNetV2} is also a fully-connected layer having one neuron.
For unseen device prediction tasks, on the other hand, the additional convolutional layers won't have the benefit of accuracy improvement.
As a result, we chose to use \emph{PerNet} architecture for the quicker inference results. 

To contrast the convergence of two network models clearly, we use the same hyper-parameters setting as \emph{PerNet}. The stand scale transformation was performed for each feature. We use Adam \cite{adam} as our optimizer for self-adaptive updating weight in each epoch. The number of epochs is 1000 with a batch size of 128.
The initial learning rate is set as 0.1, and divided by 2 every 400 epochs. It took us 4 hours to train a model for each phase of the layer on NVIDIA GTX 1080Ti.  
Details of the training process are discussed in the next section.

\section{Training the Prediction Models}
\label{section:data_collection}
In this section, we discuss how we actually extract both the software and hardware features to build the proposed performance models. 
Section \ref{section:Features} describes the features which may impact the performance in a GPU-based system. 
The features are extracted via our benchmark tool, which attempts to cover the software/hardware design space by generating a set of simple workloads (microbenchmarks) and measure their execution time on a variety of platforms. Since each microbenchmark takes at least seconds to acquire stable, accurate measurements, it is impossible to cover the entire design space with brute force, which requires over $10^{14}$ microbenchmark runs. Thus, we discuss the trade-off between time-cost and model accuracy and how to sample the design space wisely to acquire a reasonably representative training data set in practice in
Section \ref{section:Benchmark}.
It took us two weeks to obtain a training data set of 100,000 samples for each type of layer, and the prediction models described in the previous section can be built within a day with the data set.

\subsection{Features and Design Space}
\label{section:Features}
Table \ref{tab:data_description} shows the software and hardware features
used for training our prediction models.
The first 12 software features are derived from TensorFlow 1.13 APIs, including \emph{tensorflow.layers.con2d}, \emph{tensorflow.layers.max\_pooling2d}, and \emph{tensorflow.layers.dense}. 
The next 4 software features indicate the gradients descent algorithms, i.e. SGD, Adam, RMSprop, and AdaGrad, chosen for training. 
There are 5 hardware features extracted from system configurations and the architecture specifications. 
With a total of 21 features, we are able to predict the inference or training performance for most CNN's on desktop computers and servers with or without GPU's.  
This assists DL designers and users in selecting CNNs based on the system configuration.  
As the resulted model may predict the performance impact from hardware changes such as increasing the clock rate, adding more GPU cores, etc., it is of interests to system designers as well.


\subsection{Preparation of the Training Data Set} 
\label{section:Benchmark}
We develop a tool to obtain the training data set automatically. 
The tool first generates a set of the microbenchmarks by varying the features randomly and then performs the microbenchmarks on the target platforms in parallel. Finally, the time intervals measured from the Tensorflow Profiler are analyzed to extract the preprocessing time ($T_{pre}$), the execution time ($T_{exe}$), and the post-processing time ($T_{post}$), as mentioned in the previous section. 
The following describes the two main steps:

\begin{itemize}
\item
\emph{Profile Collection with Microbenchmarks:}
According to Table \ref{tab:data_description}, in inference time prediction job, there are out of $7.33 * 10^{14}$, $7.33 * 10^{10}$, and $2.14 * 10^{9}$ possibility combinations of convolutional, pooling, dense layers, respectively. While in training time prediction job, due to additional optimizer features, the maximum of possible combination, convolutional layer, is over $2.93 *10^{15}$. The number of combinations is quite huge, so we believe that it can cover most of the layers chosen for various neural networks. It is impossible to sample all combinations of these features, so we use random uniform distribution to generate a set of representative microbechmarks. 
Based on our previous experiences \cite{perfnet},
the accuracy grows as the number of samples increases, but reaches its top at ~100,000 samples.
In our experimental setup, it took us over two weeks to profile 100,000 microbenchmarks on five different systems (NVIDIA GTX1080Ti, P1000, P2000, P4000, and P5000) to train the inference models for each type of layer.
To predict the training time, we need to profile both forward and backward propagations, and thus the profiling process would take almost three times longer if the number of samples are the same. Thus, we reduce the number of samples to 30,000 to keep the profiling process within the two weeks period for the training time predictor.

\item
\emph{Data Extraction:}
To extract the time intervals from the profile data, we developed a preprocessing parser to read the reports from the Tensorflow Profiler and detect the outlier data. 
The microbenchmarks may be interfered by background processes running on the same system. The profiler may also report inaccurate timestamps occasionally. Thus, it is important to clean up the outlier data before training the prediction models.
Then, we utilize our proposed method in Section \ref{section:methodology} to split the timeline report into three phases ($T_{pre}$, $T_{exe}$, $T_{post}$) on both forward propagation and backward propagation. 
In our experiments, 80\% of the 100,000 samples of forward propagation time data collected on a system are used to train the inference time prediction model for the same system, while the remains 20\% of samples are used as the test data set for evaluating the predicted model. 
Finally, in order to predict the performance impact due to hardware changes, we train a prediction model by combining the training data set collected from P1000, P2000, and P4000. The combined training data set contains 300,000 samples, the test data set is 100,000 samples collected from P5000, and the training process takes several days for our GTX1080Ti-based computer to produce a good predictor.
\end{itemize}

\begin{table*}
\begin{center}
\caption{{\upshape Description of features.}}
\label{tab:data_description} 
\begin{tabular}{p{1.5cm}p{10cm}ccc}
Name     & Description & Type & Range & Scenario$^{\mathrm{1}}$ \\ \hline
Batch Size     & The number of parallel processed in one iteration. & Software & 1-64 & All  \\ 
Matrix Size    & The dimensions of the input data. & Software & 1-512 & Conv2D, Pooling\\
Kernel Size    & The size of the filter applied to the input data.  & Software &  1-7 & Conv2D \\
Channel In     & The number of channels in the input data.  & Software & 1-9999 & Conv2D, Pooling \\
Channel Out    & The number of channels in the output data.  & Software & 1-9999 & Conv2D \\
Strides        & The amount of the window shifts for each dimension of the input data with kernels.  & Software & 1-4 & Conv2D, Pooling \\
Padding        & The number for preserving the original size of the image while the filter scans each pixel and makes the input data smaller. 0: Valid, 1: Same.  & Software & 0-1 & Conv2D, Pooling \\
Activate Function& The number for representing what activation function is used. 0: Without an activate function. 1: Relu activate function.  & Software & 0-1 & All\\
Bias             & The boolean number for utilizing an additional intercept on training input data.  & Software & 0-1 & Conv2D, Dense\\
Dimension Input  & The number of outputs from the previous layer.  & Software & 1-4096 & Dense\\
Dimension Output & The number of outputs of the layer.  & Software & 1-4096 & Dense\\
Pooling Size     & The windows size factors for scaling down the input data.  & Software & 1-7 & Pooling\\
SGD              & Stochastic gradient descent (SGD) is enabled or not during the training process. 
& Software & 0-1 & All-Training \\
AdaGrad          & Adaptive gradient algorithm (AdaGrad) is enabled or not during the training process. 
& Software  & 0-1 & All-Training\\
RMSprop          & Root mean square propagation (RMSprop) is enabled or not during the training process.
& Software & 0-1 & All-Training \\
Adam             & Adaptive moment estimation (Adam) is enabled or not during the training process.
& Software & 0-1 & All-Training \\

Basic Clock & The clock cycles per second for a DL accelerator (MHz).  & Hardware & 1076-1607 & All-Unseen \\
CUDA Cores        & The number of parallel processors responsible for data processing that is fed into and out of a DL card and performing graphics calculation asynchronously for user. & Hardware & 640-3584 & All-Unseen \\
Memory Clock & The number of clock cycles per second. Having higher memory clock allows a DL card to process the memory tasks quicker (MHz). & Hardware & 1127-1901 & All-Unseen \\ 
Memory Bandwidth & The bandwidth for the DL accelerator to access its memory (GB/s). & Hardware & 80.19-484.4 & All-Unseen \\
Peak TFLOPs & The highest 32-bit floating point operations per second for the DL accelerator based on the manufacturer's specifications. & Hardware & 1.894-11.34 & All-Unseen \\ \hline
\multicolumn{5}{p{18cm}} {$^{\mathrm{1}}$ In this scenario column, All means the feature would be used in all types of the layer. Conv2D, Pooling, and Dense stand for the feature would be used in respective type of layer for the all predicted jobs, including inference, training, and unseen device. Similarly, All-Training means the feature would only be used for training jobs in all types of the layer, and All-Unseen is only for unseen prediction jobs.}
\end{tabular}
\end{center}
\end{table*}

\section{Performance Evaluation}
\label{section:performance_evaluation}
In this section, we evaluate the proposed modeling methods via experiments. 
Section \ref{subsection:hard_devices} describes our experiment setup, including the hardware specifications and the metrics for comparing the performance of the proposed methods against the previous works. 
The remaining subsections evaluate the prediction accuracy of the proposed methods in five categories: inference time for individual layers, training time for individual layers, inference time for full networks, training time for full networks, and inference time for unseen devices.

\subsection{Hardware Devices and Performance Metrics}
\label{subsection:hard_devices}
The hardware devices used in our experiments and their features are listed in Table \ref{tab:gpu_devices_description}. 
All these GPU devices are based on NIVIDIA's Pascal microarchitecture. Each of them is plugged into a 16-lane PCIe 3.0 slot in an individual machine. Each machine contains an Intel i7-7700 quad-core CPU. In fact, the choice of the CPU has very little effects the prediction model, since this study focuses on the execution of the DNN on the GPU devices.

\begin{table}
\begin{center}
\caption{{\upshape Specifications of hardware.}} 
\label{tab:gpu_devices_description} 
\scalebox{0.85}{
\begin{tabular}{|c|c|c|c|c|c|}
\hline
NVIDIA & Basic Clock & CUDA  & Memory      & Memory Band-    & Peak  \\ 
Device & (MHz)       & Cores & Clock (MHz) & width (GB/s)    & TFLOPS \\ \hline
P1000     & 1266 &  640 & 1253 & 80.19 & 1.894 \\ \hline
P2000     & 1076 & 1024 & 1752 & 140.2 & 3.031 \\ \hline
P4000     & 1202 & 1792 & 1901 & 243.3 & 5.304 \\ \hline
P5000     & 1607 & 2560 & 1127 & 288.5 & 8.873 \\ \hline
GTX1080Ti & 1481 & 3584 & 1376 & 484.4 & 11.34 \\ \hline
\end{tabular}
}
\end{center}
\end{table}

For performance evaluation, we use coefficient of determination ($R^2$) to determine the quality of prediction. The value of $R^2$ is always between 0.0 and 1.0. $R^2$ over 0.9 means that the model is highly confident.
To further analyze the accuracy of the prediction models, we use mean absolute percentage error (MAPE), mean absolute error (MAE) and root mean squared error (RMSE) as performance metrics.
MAPE calculates the prediction error by averaging the error percentages in the individual predictions.
Unlike MAPE, MAE calculates the arithmetic average of the absolute errors, which can be dominated by the errors in predicting large absolute values. 
RMSE is often used in statistics, which calculates the standard deviation of the errors. 

\subsection{Inference Time for Individual Layers}
\label{subsection:inference_individual}
\begin{figure}
\centering
\includegraphics[width=3.5in]{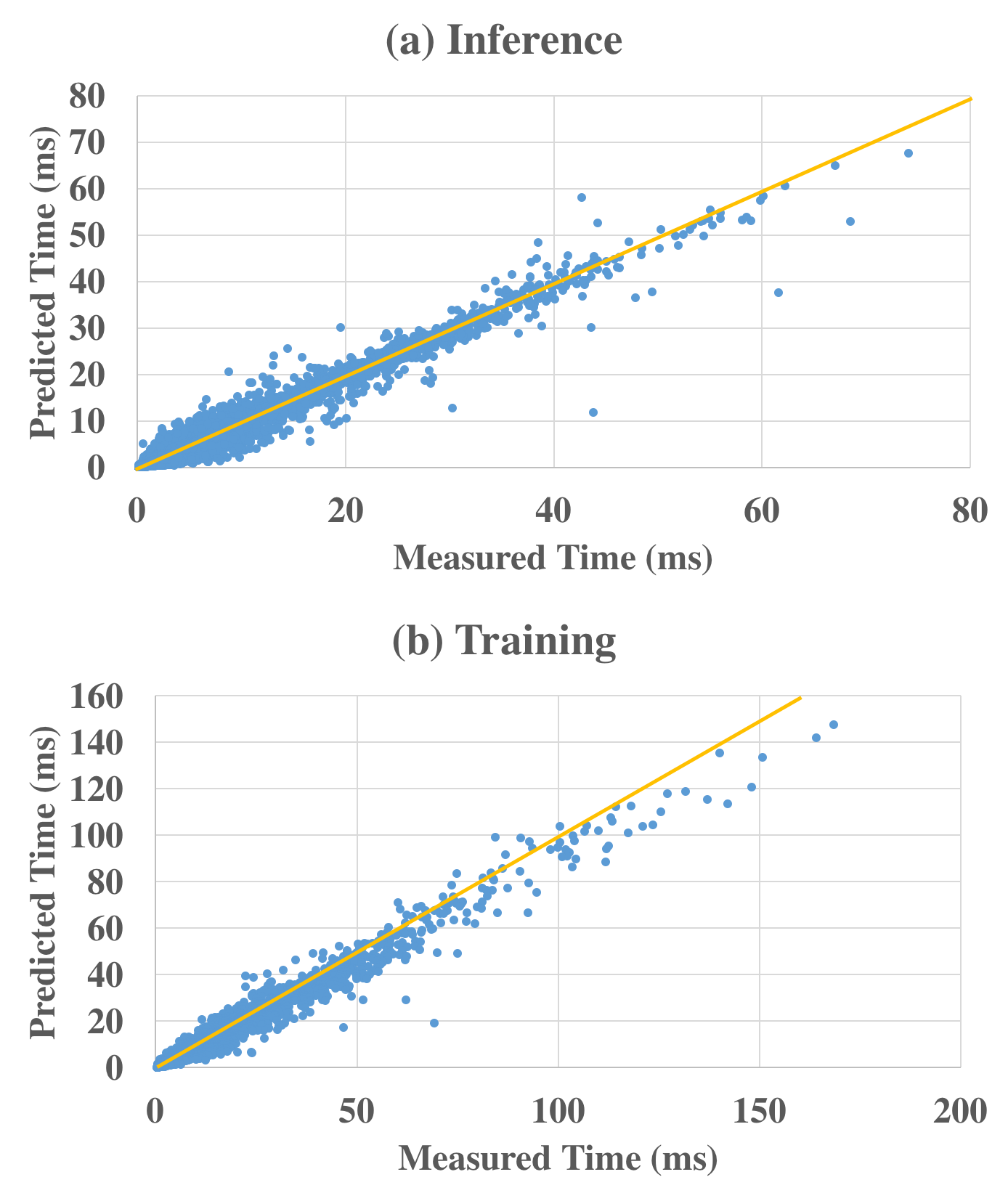}
\caption{Predicted time versus measured time (a) Inference Time (b) Training Time for execution phase of convolutional layer on NVIDIA GTX 1080Ti.}
\label {fig:spread}
\end{figure}

In this section, we predict the inference time for individual layers and evaluate the quality of predictions in various test cases. 
As shown in Table \ref{tab:inf_all_1080Ti}, the $R^2$ values are mostly over 0.95 which indicates high prediction quality for \emph{PerNetV2}, except for the dense layers.
The MAPE of each phase is under 14\%, which helps remove the worries of over-fitting.
The largest RMSE and MAE for inference time prediction are 0.92ms and 0.362ms, respectively, which occur in the prediction of the execution phase of the convolutional layers. 
As shown in Figure~\ref{fig:spread}(a), we depict the comparison between predicted and measured inference time in detail for execution phase of convolutional layers on NVIDIA GTX 1080Ti.
Table \ref{tab:inf_rmse_mape_p} gives a complete list of the RMSE's and MAPE's 
for predicting the time spent in the execution phase on the P-Series GPU's.
The RMSE's for slower devices, such as the P1000, tend to be higher since it takes longer for the slower devices to execute the same workload, but the percentage errors, i.e. MAPE's, are similar across different devices.

\begin{table}
\begin{center}
\caption{{ \emph{PerfNetV2}\upshape: Accuracy of inference/training predicted result on NVIDIA-1080Ti.}}
\label{tab:inf_all_1080Ti}
\scalebox{0.8}{
\begin{tabular}{|c|c|c|c|c|c|c|}
\hline
Types    & Layers & Phases  & $R^2$ & MAPE(\%) & RMSE(ms) & MAE(ms) \\ \hline
\multirow{9}{*}{Inference}
    & \multirow{3}{*}{convolutional} 
        & preprocess  & 0.99 & 2.387  & 0.233 & 0.148 \\ \cline{3-7} 
    &   & execution   & 0.97 & 13.056 & 0.92  & 0.362 \\ \cline{3-7} 
    &   & postprocess & 0.99 & 3.754  & 0.108 & 0.049 \\ \cline{2-7} 
    & \multirow{3}{*}{pooling}
        & preprocess  & 0.99 & 2.051  & 0.214 & 0.132 \\ \cline{3-7} 
    &   & execution   & 0.99 & 7.847  & 0.364 & 0.127 \\ \cline{3-7} 
    &   & postprocess & 0.99 & 3.087  & 0.102 & 0.048 \\ \cline{2-7} 
    & \multirow{3}{*}{dense}
        & preprocess  & 0.96 & 6.543  & 0.037 & 0.025 \\ \cline{3-7} 
    &   & execution   & 0.98 & 4.529  & 0.068 & 0.036 \\ \cline{3-7} 
    &   & postprocess & 0.72 & 13.879 & 0.011 & 0.009 \\ \cline{2-7} \hline 
\multirow{9}{*}{Training}
    & \multirow{3}{*}{convolutional} 
        & preprocess  & 0.99 & 5.111  & 0.607 & 0.371 \\ \cline{3-7} 
    &   & execution   & 0.97 & 14.229 & 2.561 & 1.108 \\ \cline{3-7} 
    &   & postprocess & 0.99 & 3.754  & 0.108 & 0.049 \\ \cline{2-7} 
    & \multirow{3}{*}{pooling}
        & preprocess  & 0.99 & 5.187  & 0.730 & 0.419 \\ \cline{3-7} 
    &   & execution   & 0.99 & 7.073  & 0.453 & 0.146 \\ \cline{3-7} 
    &   & postprocess & 0.99 & 7.779  & 0.072 & 0.035 \\ \cline{2-7} 
    & \multirow{3}{*}{dense}
        & preprocess  & 0.87 & 8.017  & 0.080 & 0.047 \\ \cline{3-7} 
    &   & execution   & 0.70 & 22.737 & 5.503 & 3.455 \\ \cline{3-7} 
    &   & postprocess & 0.78 & 18.012 & 0.018 & 0.011 \\ \cline{2-7} \hline 
\end{tabular}}
\end{center}
\end{table}

\begin{table*}[]
\begin{center}
\caption{{\upshape Predicted inference results of the execution phase in 3 major layers on hardware.}}
\label{tab:inf_rmse_mape_p}
\begin{tabular}{lllllllllll}
                             & \textbf{(a) convolutional}       &                                &                       &                             & \textbf{(b) pooling}           &                                &                       &                             & \textbf{(c) dense}             &                                \\ \cline{1-3} \cline{5-7} \cline{9-11} 
\multicolumn{1}{|l|}{Device} & \multicolumn{1}{c|}{RMSE (ms)} & \multicolumn{1}{c|}{MAPE (\%)} & \multicolumn{1}{c|}{} & \multicolumn{1}{c|}{Device} & \multicolumn{1}{c|}{RMSE (ms)} & \multicolumn{1}{c|}{MAPE (\%)} & \multicolumn{1}{l|}{} & \multicolumn{1}{c|}{Device} & \multicolumn{1}{c|}{RMSE (ms)} & \multicolumn{1}{c|}{MAPE (\%)} \\ \cline{1-3} \cline{5-7} \cline{9-11} 
\multicolumn{1}{|c|}{P1000}  & \multicolumn{1}{c|}{5.681}     & \multicolumn{1}{c|}{15.805}    & \multicolumn{1}{l|}{} & \multicolumn{1}{c|}{P1000}  & \multicolumn{1}{c|}{1.162}     & \multicolumn{1}{c|}{10.898}    & \multicolumn{1}{l|}{} & \multicolumn{1}{c|}{P1000}  & \multicolumn{1}{c|}{0.439}     & \multicolumn{1}{c|}{6.002}     \\ \cline{1-3} \cline{5-7} \cline{9-11} 
\multicolumn{1}{|c|}{P2000}  & \multicolumn{1}{c|}{4.666}     & \multicolumn{1}{c|}{15.377}    & \multicolumn{1}{l|}{} & \multicolumn{1}{c|}{P2000}  & \multicolumn{1}{c|}{0.763}     & \multicolumn{1}{c|}{9.445}     & \multicolumn{1}{l|}{} & \multicolumn{1}{c|}{P2000}  & \multicolumn{1}{c|}{0.334}     & \multicolumn{1}{c|}{4.989}     \\ \cline{1-3} \cline{5-7} \cline{9-11} 
\multicolumn{1}{|c|}{P4000}  & \multicolumn{1}{c|}{1.995}     & \multicolumn{1}{c|}{16.004}    & \multicolumn{1}{l|}{} & \multicolumn{1}{c|}{P4000}  & \multicolumn{1}{c|}{0.564}     & \multicolumn{1}{c|}{12.327}    & \multicolumn{1}{l|}{} & \multicolumn{1}{c|}{P4000}  & \multicolumn{1}{c|}{0.185}     & \multicolumn{1}{c|}{5.994}     \\ \cline{1-3} \cline{5-7} \cline{9-11} 
\multicolumn{1}{|c|}{P5000}  & \multicolumn{1}{c|}{1.970}     & \multicolumn{1}{c|}{14.840}    & \multicolumn{1}{l|}{} & \multicolumn{1}{c|}{P5000}  & \multicolumn{1}{c|}{0.578}     & \multicolumn{1}{c|}{10.581}    & \multicolumn{1}{l|}{} & \multicolumn{1}{c|}{P5000}  & \multicolumn{1}{c|}{0.130}     & \multicolumn{1}{c|}{5.768}     \\ \cline{1-3} \cline{5-7} \cline{9-11} 
\end{tabular}
\end{center}
\end{table*}

\subsection{Training Time for Individual Layers}
\label{subsection:training_individual}
Unlike \emph{PerfNet}, \emph{PerfNetV2} is capable of predicting the time needed to perform the individual phases during the training process.
As shown in Table \ref{tab:inf_all_1080Ti}, the $R^2$ values for the case of training are mostly good (over 0.97) except for the dense layers.
The RMSE's and MAE's are generally larger for the case of training since each phase contains both forward propagation and backward propagation. 
The MAPE's for the case of training are up to 22.7\%, which occurs for the dense layers.
However, as shown in Table \ref{tab:train_rmse_mape_p}, when \emph{PerfNetV2} is used to predict the training time on the P-Series devices, the MAPE's for the dense layers are surprisingly low. While the accuracy is still acceptable for the case of training, it is less stable than the case of inference, and it is to be further investigated in our future work.
In addition, we also illustrate the comparison between predicted and measured training time for execution phase of convolutional layers on NVIDIA GTX 1080Ti, as shown in Figure~\ref{fig:spread}(b).



\begin{table*}[]
\begin{center}
\caption{{\upshape Predicted training results of the execution phase in 3 major layers on hardware.}}
\label{tab:train_rmse_mape_p}
\begin{tabular}{lllllllllll}
                             & \textbf{(a) convolutional}       &                                &                       &                             & \textbf{(b) pooling}           &                                &                       &                             & \textbf{(c) dense}             &                                \\ \cline{1-3} \cline{5-7} \cline{9-11} 
\multicolumn{1}{|c|}{Device} & \multicolumn{1}{c|}{RMSE (ms)} & \multicolumn{1}{c|}{MAPE (\%)} & \multicolumn{1}{c|}{} & \multicolumn{1}{c|}{Device} & \multicolumn{1}{c|}{RMSE (ms)} & \multicolumn{1}{c|}{MAPE (\%)} & \multicolumn{1}{c|}{} & \multicolumn{1}{c|}{Device} & \multicolumn{1}{c|}{RMSE (ms)} & \multicolumn{1}{c|}{MAPE (\%)} \\ \cline{1-3} \cline{5-7} \cline{9-11} 
\multicolumn{1}{|c|}{P1000}  & \multicolumn{1}{c|}{15.443}    & \multicolumn{1}{c|}{13.588}    & \multicolumn{1}{c|}{} & \multicolumn{1}{c|}{P1000}  & \multicolumn{1}{c|}{2.744}     & \multicolumn{1}{c|}{7.650}     & \multicolumn{1}{c|}{} & \multicolumn{1}{c|}{P1000}  & \multicolumn{1}{c|}{0.739}     & \multicolumn{1}{c|}{2.430}     \\ \cline{1-3} \cline{5-7} \cline{9-11} 
\multicolumn{1}{|c|}{P2000}  & \multicolumn{1}{c|}{12.670}    & \multicolumn{1}{c|}{12.010}    & \multicolumn{1}{c|}{} & \multicolumn{1}{c|}{P2000}  & \multicolumn{1}{c|}{3.184}     & \multicolumn{1}{c|}{9.088}     & \multicolumn{1}{c|}{} & \multicolumn{1}{c|}{P2000}  & \multicolumn{1}{c|}{0.591}     & \multicolumn{1}{c|}{2.373}     \\ \cline{1-3} \cline{5-7} \cline{9-11} 
\multicolumn{1}{|c|}{P4000}  & \multicolumn{1}{c|}{5.636}     & \multicolumn{1}{c|}{9.172}     & \multicolumn{1}{c|}{} & \multicolumn{1}{c|}{P4000}  & \multicolumn{1}{c|}{2.516}     & \multicolumn{1}{c|}{8.444}     & \multicolumn{1}{c|}{} & \multicolumn{1}{c|}{P4000}  & \multicolumn{1}{c|}{0.386}     & \multicolumn{1}{c|}{3.578}     \\ \cline{1-3} \cline{5-7} \cline{9-11} 
\multicolumn{1}{|c|}{P5000}  & \multicolumn{1}{c|}{5.797}     & \multicolumn{1}{c|}{8.960}     & \multicolumn{1}{c|}{} & \multicolumn{1}{c|}{P5000}  & \multicolumn{1}{c|}{2.035}     & \multicolumn{1}{c|}{7.258}     & \multicolumn{1}{c|}{} & \multicolumn{1}{c|}{P5000}  & \multicolumn{1}{c|}{0.320}     & \multicolumn{1}{c|}{2.746}     \\ \cline{1-3} \cline{5-7} \cline{9-11}
\end{tabular}
\end{center}
\end{table*}

\subsection{Inference Time for Full Networks}
\label{subsection:prediction_full_networks_inference_time}
\begin{figure*}
\centering
\includegraphics[width=7.25in]{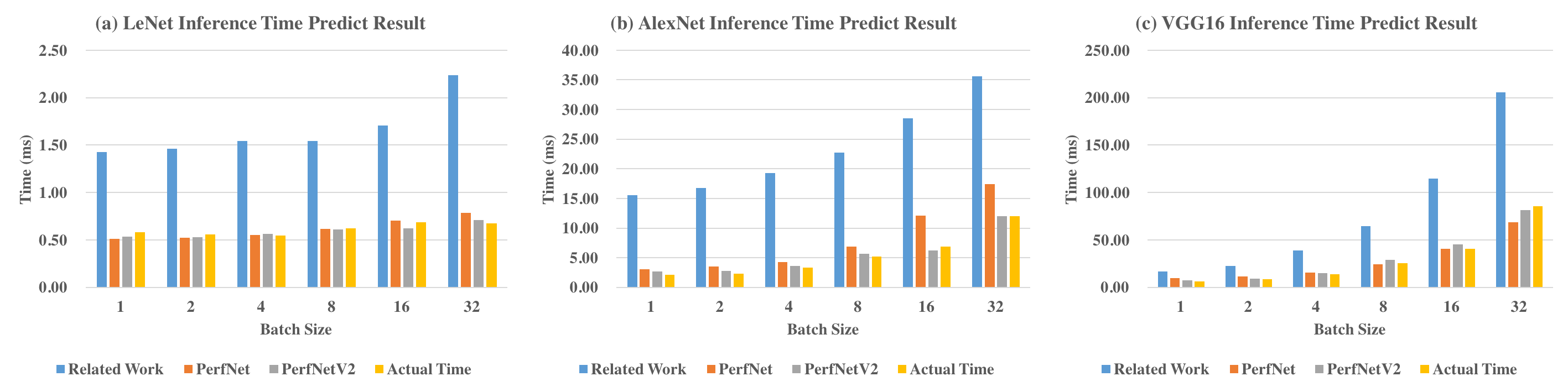}
\caption{Predicted full model results comparison among (a) LeNet (b) AlexNet (c) VGG16 on NVDIA GTX1080Ti.}
\label {fig:1080Ti_full_model_compare}
\end{figure*}

In this section, we predict the inference time of the three well-known full deep neural networks (DNNs) on NVIDIA GTX1080Ti, including LeNet, AlexNet, and VGG16, with the related work \cite{daniel2018}, \emph{PerfNet} and \emph{PerfNetV2}. The prediction results for various batch sizes are compared to the actual execution time, as shown in Figure \ref{fig:1080Ti_full_model_compare}.
The effects of varying the batch size is difficult to predict with analytic models as it increases the GPU utilization at the cost of memory usages and cache spaces. 
Obviously, the prediction from \cite{daniel2018} is less accurate than \emph{PerfNet} and \emph{PerfNetV2}, due to its coarse-grain model design. 
\emph{PerfNetV2} is more accurate than the other two predictors in most cases.
Overall, the MAPE for \emph{PerfNetV2} is 13.1\%, which is a significant improvement over \emph{PerfNet}, whose MAPE is 24.04\%.


\begin{figure*}
\centering
\includegraphics[width=7.25in]{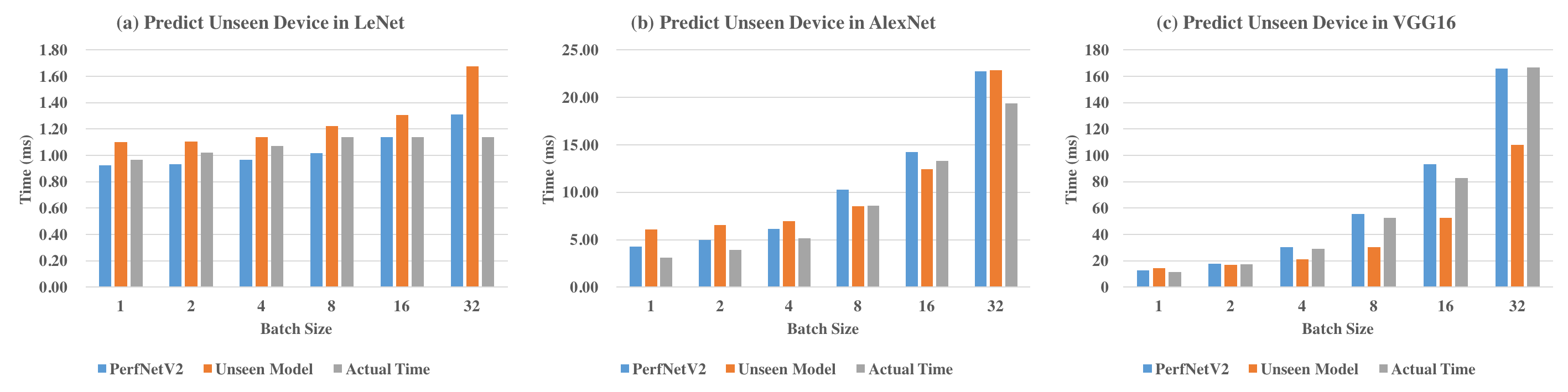}
\caption{Unseen device prediction result on (a) LeNet (b) AlexNet (c) VGG16.}
\label {fig:predict_unseen_device}
\end{figure*}

\begin{figure}
\centering
\includegraphics[width=3.6in]{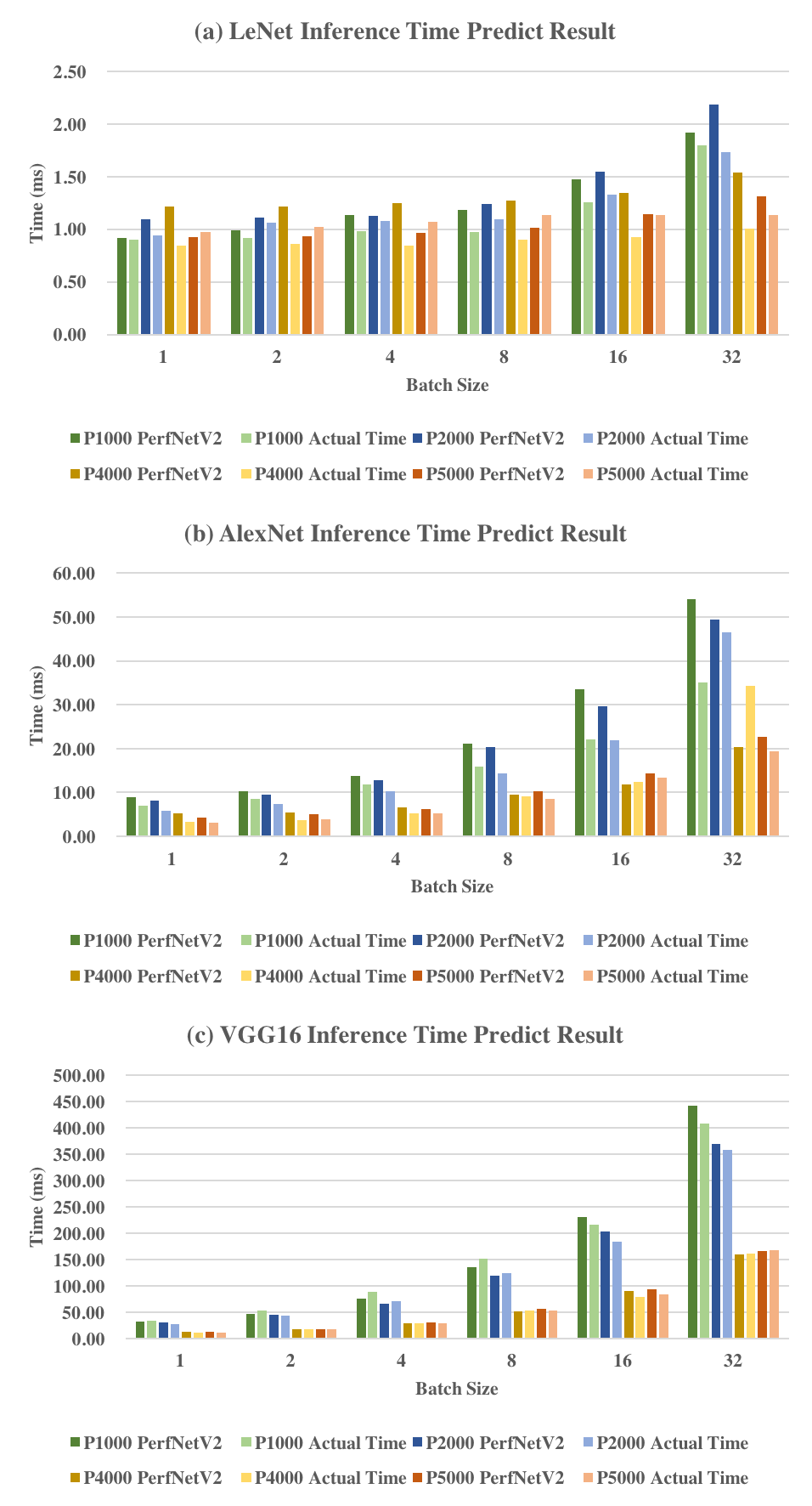}
\caption{Inference Time Prediction: Comparison between \emph{PerfNetV2} and actual inference time for (a) LeNet, (b) AlexNet and (c) VGG16 on all experimental hardware devices.}
\label {fig:full_model_inference_all}
\end{figure}

In Figure \ref{fig:full_model_inference_all}, we show that the inference time predicted by \emph{PerfNetV2} for performing the same three DNN's with different batch sizes on the four P-Series GPU devices.
The results are of interests to system designers in choosing a proper GPU for a given DNN. 
Based on Table \ref{tab:gpu_devices_description}, the P5000 has the highest TFLOPS and is supposed to outperform the other GPU devices. 
However, with small batch sizes, a high-end GPU device does not necessarily outperform a low-end device. Actually, according to our measurements, the P5000 may not perform better than the P4000 does even with a large batch size. 
It is very difficult to explain this phenomena as it involves complicated hardware-software interactions.
With our machine-learning approach, we are able to obtain a working performance model to help system designers.

The results for predicting the performance of LeNet are shown in Figure~\ref{fig:full_model_inference_all}(a), where our model accurately predicts the performance on P5000 at batch size 16 with 0.25\% relative error, and the average relative error for all batch sizes on these four systems is 19.79\%.
In Figure~\ref{fig:full_model_inference_all}(b) we depict the evaluation results for AlexNet. The average relative errors for all batch sizes on these devices is 28.75\%,
which is not as good as the cases with LeNet and VGG16. 
Figure~\ref{fig:full_model_inference_all}(c) shows the prediction results for VGG16, which is quite accurate for across different batch sizes with an average of 7.37\% relative error on the P5000.  
The overall average relative prediction error for all batch sizes and all the 3 DNNs on the four GPU devices is 18.63\%.

\subsection{Training Time for Full Networks}
\label{subsection:prediction_full_networks_training_time}
\begin{figure}
\centering
\includegraphics[width=3.75in]{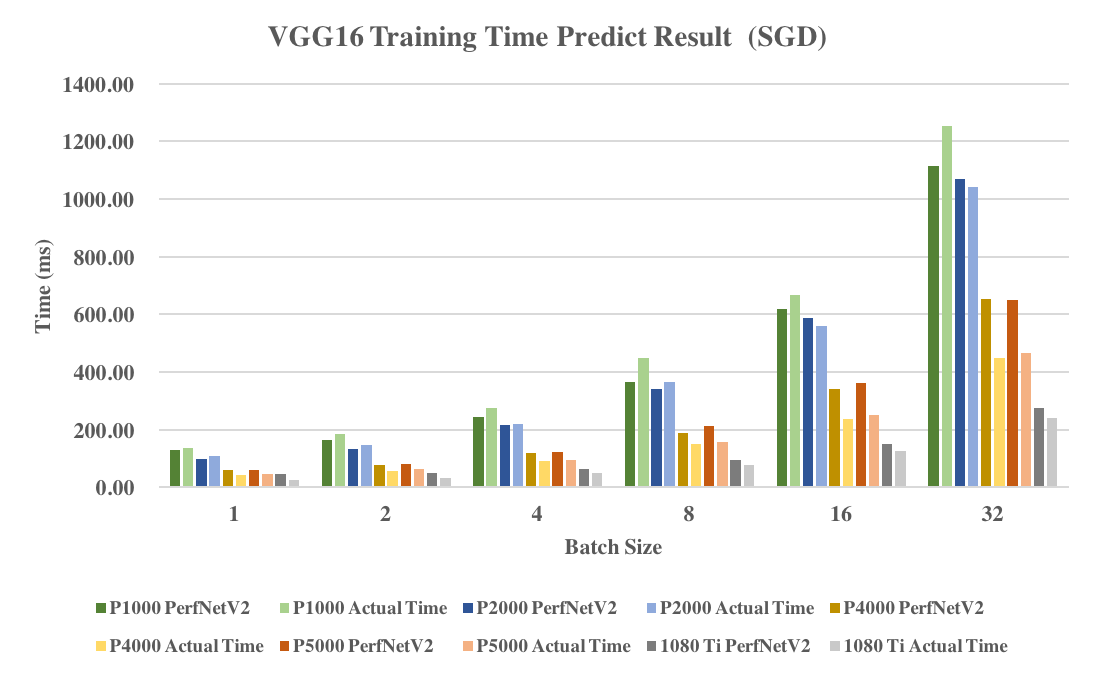}
\caption{Training Time Prediction: Comparison between \emph{PerfNetV2} and actual training time for VGG16 on all experimental hardware devices by using SGD as the optimizer.}
\label {fig:vgg16_traing_all}
\end{figure}

Figure \ref{fig:vgg16_traing_all} shows the predicted time and the actual time for all the five GPU devices to execute one forward and backward propagation round in the case of training a full VGG16 DNN with the SGD optimizer.
The average relative error for all batch sizes and all the three DNN's on all the five GPU devices is 21.65\%, which is still acceptable in practice.


\subsection{Inference Time of Unseen Devices}
\label{subsection:prediction_full_networks_inference_time_unseen}
Given an unseen device whose performance data have not been included in the training data set, our model can still give performance estimates. For example, suppose we have built a prediction model based on the performance data collected from P1000, P2000 and P4000, and would like to estimate the performance of P5000, with this so called \emph{Unseen Model}.
Figure \ref{fig:predict_unseen_device} compares the predicted inference time produced by \emph{PerfNetV2} and the aforementioned \emph{Unseen Model}, to the actual time of executing three DNN's on the P5000.
Overall, the average relative errors for \emph{PerfNetV2} and \emph{Unseen Model} are 12.05\% and 27.38\%, respectively. 
While \emph{Unseen Model} is less accurate, it still provides useful estimates for system designers.


\section{Conclusion and Future Work}
\label{section:conclusion_futurework}

In this paper, we propose platform-aware performance modeling techniques to improve the accuracy of the previously proposed methods by considering the interactions between the host and the GPU and separating a neural network operation into three phases. 
In order to build the proposed fine-grain performance models, we develop a set of tools to automatically collect performance-related data and extract features on a variety of platforms. 
Our experimental results suggest that the proposed platform-aware performance model delivers practical estimates which are useful to application developers and system designers in choosing suitable neural networks and/or proper platforms during the early-stage planning.

In the future, 
we would like to bring more contribution to the community by extending our approach to model more hardware devices, such as Intel's VPU \cite{vpu}, and cover more deep learning frameworks, such as OpenVINO \cite{openVino} and TensorRT \cite{tensorRT}.
It is possible to further improve the accuracy of our performance models with feature transformation and innovative neural network architectures for the regression tasks.
Furthermore, we intend to model the performance of high-performance deep learning systems, such as a distributed multi-GPU computing cluster, by additional modeling of the inter-GPU communications. 


\end{document}